\documentclass[runningheads]{llncs}
\usepackage{graphicx}
\usepackage{algorithm}
\usepackage{algorithmic}
\usepackage{booktabs}  
\usepackage{threeparttable}  
\begin{document}
%
\title{Important Attribute Identification in Knowledge Graph}
\titlerunning{Important Attribute Identification}
%
%


\author{Shengjie Sun,
Dong Yang,
Hongchun Zhang,
Yanxu Chen,
ChaoWei,
Xiaonan Meng,
Yi Hu}
\authorrunning{.}
%
\institute{Alibaba Group\\ 
Hangzhou, China\\
\email{\{shengjie.ssj,
	dong.yd,
	hongchun.zhc,
	yanxu.cyx,
	weichao.wc,
	xiaonan.mengxn,
	erwin.huy
	\}@alibaba-inc.com}}

%
\maketitle              
\begin{abstract}
The knowledge graph(KG) composed of entities with their descriptions and attributes, and relationship between entities, is finding more and more application scenarios in various natural language processing tasks. In a typical knowledge graph like Wikidata, entities usually have a large number of attributes, but it is difficult to know which ones are important. The importance of attributes can be a valuable piece of information in various applications spanning from information retrieval to natural language generation. In this paper, we propose a general method of using external user generated text data to evaluate the relative importance of an entity's attributes. To be more specific, we use the word/sub-word embedding techniques to match the external textual data back to entities' attribute name and values and rank the attributes by their matching cohesiveness. To our best knowledge, this is the first work of applying vector based semantic matching to important attribute identification, and our method outperforms the previous traditional methods. We also apply the outcome of the detected important attributes to a language generation task; compared with previous generated text, the new method generates much more customized and informative messages. 
\keywords{Knowledge graph  \and Attribute importance \and Important attribute identification.}
\end{abstract}
\section{Introduction}
\subsection{The problem we solve in this paper}

Knowledge graph(KG) has been proposed for several years and its most prominent application is in web search, for example, Google search triggers a certain entity card when a user's query matches or mentions an entity based on some statistical model.  The core potential of a knowledge graph is about its capability of reasoning and inferring, and we have not seen revolutionary breakthrough in such areas yet.   One main obstacle is obviously the lack of sufficient knowledge graph data, including entities, entities' descriptions, entities' attributes, and relationship between entities. A full functional knowledge graph supporting general purposed reasoning and inference might still require long years of the community's innovation and hardworking. On the other hand, many less demanding applications have great potential benefiting from the availability of information from the knowledge graph, such as query understanding and document understanding in information retrieval/search engines, simple inference in question answering systems, and easy reasoning in domain-limited decision support tools.  Not only academy, but also industry companies have been heavily investing in knowledge graphs, such as Google's knowledge graph, Amazon's product graph, Facebook's Graph API, IBM's Watson, and Microsoft's Satori etc.

In the existing knowledge graph, such as Wikidata and DBpedia, usually attributes do not have order or priorities, and we don't know which attributes are  more important and of more interest to users. Such importance score of attributes is a vital piece of information in many applications of knowledge graph. The most important application is the triggered entity card in search engine when a customer's query gets hit for an entity. An entity usually has a large amount of attributes, but an entity card has limited space and can only show the most significant information; attribute importance's presence can make the displaying of an entity card easy to implement. Attribute importance also has great potential of playing a significant role in search engine, how to decide the matching score between the query and attribute values. If the query matches a very important attribute, and the relevance contribution from such a match should be higher than matching an ignorable attribute.  Another application is in e-commerce communications, and one buyer initiates a communication cycle with a seller by sending a product enquiry. Writing the enquiry on a mobile phone is inconvenient and automatic composing assistance has great potential of improving customer experience by alleviating the writing burden. In the product enquiry, customers need to specify their requirements  and ask questions about products, and their requirements and questions are usually about the most important attributes of the products. If we can identify out important attributes of products, we can help customers to draft the enquiry automatically to reduce their input time. 

\subsection{Related Research}\label{sec:related}

Many proposed approaches formulate the entity attribute ranking problem as a post processing step of automated attribute-value extraction. In \cite{pacsca2010role,pasca2007you,pacsca2007role}, Pasca et al. firstly extract potential class-attribute pairs using linguistically motivated patterns from unstructured text  including query logs and query sessions, and then score the attributes using the Bayes model. In \cite{rai2012identifying}, Rahul Rai proposed to identify product attributes from customer online reviews using part-of-speech(POS) tagging patterns, and to evaluate their importance with several different frequency metrics. In \cite{lee2013attribute}, Lee et al. developed a system to extract concept-attribute pairs from multiple data sources, such as Probase, general web documents, query logs and external knowledge base, and aggregate the weights from different sources into one consistent typicality score using a Ranking SVM model. Those approaches typically suffer from the poor quality of the pattern rules, and the ranking process is used to identify relatively more precise attributes from all attribute candidates.

As for an already existing knowledge graph, there is plenty of work in literature dealing with ranking entities by relevance without or with a query. In \cite{ding2005finding}, Li et al. introduced the OntoRank algorithm for ranking the importance of semantic web objects at three levels of granularity: document, terms and RDF graphs. The algorithm is based on the rational surfer model, successfully used in the Swoogle semantic web search engine. In \cite{hogan2006reconrank}, Hogan et al. presented an approach that adapted the well-known PageRank/HITS algorithms to semantic web data, which took advantage of property values to rank entities. In \cite{graves2008method,he2010xhrank}, authors also focused on ranking entities, sorting the semantic web resources based on importance, relevance and query length, and aggregating the features together with an overall ranking model.

Just a few works were designated to specifically address the problem of computing attribute rankings in a given Knowledge Graph. Ibminer \cite{mousavi2013ibminer} introduced a tool for infobox(alias of an entity card) template suggestion, which collected attributes from different sources and then sorted them by popularity based on their co-occurrences in the dataset. In \cite{vadrevu2016ranking}, using the structured knowledge base, intermediate features were computed, including the importance or popularity of each entity type, IDF computation for each  attribute on a global basis, IDF computation for entity types etc., and then the features were aggregated to train a classifier. Also, a similar approach in \cite{atzori2014ranking} was designed with more features extracted from GoogleSuggestChars data. In \cite{alientity}, Ali et al. introduced a new set of features that utilizes semantic information about entities as well as information from top-ranked documents from a general search engine. In order to experiment their approach, they collected a dataset by exploiting Wikipedia infoboxes, whose ordering of attributes reflect the collaborative effort of a large community of users, which might not be accurate.

\subsection{What we propose and what we have done}

There have been broad researches on entity detection, relationship extraction, and also missing relationship prediction. For example:  \cite[Lin etc.]{lin15relationship}, \cite[Wang etc.]{2014wangembedding} and \cite[Amit etc.]{amit2012knowledgegraph} explained how to construct a knowledge graph and how to perform representation learning on knowledge graphs. Some research has been performed on attribute extraction, such as \cite{ali2017attribute} and \cite{lee2013attribute}; the latter one is quite special that it also simultaneously computes the attribute importance. As for modeling attribute importance for an existing knowledge graph which has completed attribute extractions, we found only a few existing research, all of which used simple co-occurrences to rank entity attributes.  In reality, many knowledge graphs do not contain attribute importance information, for example, in the most famous Wikidata, a large amount of entities have many attributes, and it is difficult to know which attributes are significant and deserve more attention. In this research we focus on identifying important attributes in existing knowledge graphs. Specifically,  we propose a new method of using extra user generated data source for evaluating the attribute importance, and we use the recently proposed state-of-the-art word/sub-word embedding techniques to match the external data with the attribute definition and values from entities in knowledge graphs. And then we use the statistics obtained from the matching to compare the attribute importance. Our method has general extensibility to any knowledge graph without attribute importance. When there is a possibility of finding external textual data source, our proposed method will work, even if the external data does not exactly match the attribute textual data, since the vector embedding performs semantic matching and does not require exact string matching.

The remaining of the paper is organized as follows: Section~\ref{sec:ourmethod} explains our proposed method in detail, including what kind of external data is required, and how to process the external data, and also how to perform the semantic matching and how to rank the attributes by statistics. Section~\ref{sec:experiment} introduces our experimentations, including our experimentation setup, data introduction and experimental result compared to other methods we do not employ. Section~\ref{sec:experiment} also briefly introduces our real world application scenario in e-commerce communication.  Section~\ref{sec:futurework} draws the conclusion from our experimentations and analysis, and also we point out promising future research directions.

\section{Our proposed Method}\label{sec:ourmethod}
In this section, we will introduce our proposed method in detail. We use our application scenario to explain the logic behind the method, but the scope is not limited to our use case, and it is possible to extend to any existing knowledge graph without attribute importance information. 

\subsection{Application  Scenario}
Alibaba.com is currently the world's largest cross-border business to business(B2B) E-commerce platform and it supports 17 languages for customers from all over the world. On the website, English is the dorminant language and accounts for around 50\% of the traffic. The website has already accumulated a very large knowledge graph of products, and the entity here is the product or the product category; and every entity has lots of information such as the entity name, images and many attributes without ordering information. The entities are also connected by taxonomy structure and similar products usually belong to the same category/sub-category. 

\begin{figure}
	\includegraphics[width=\textwidth]{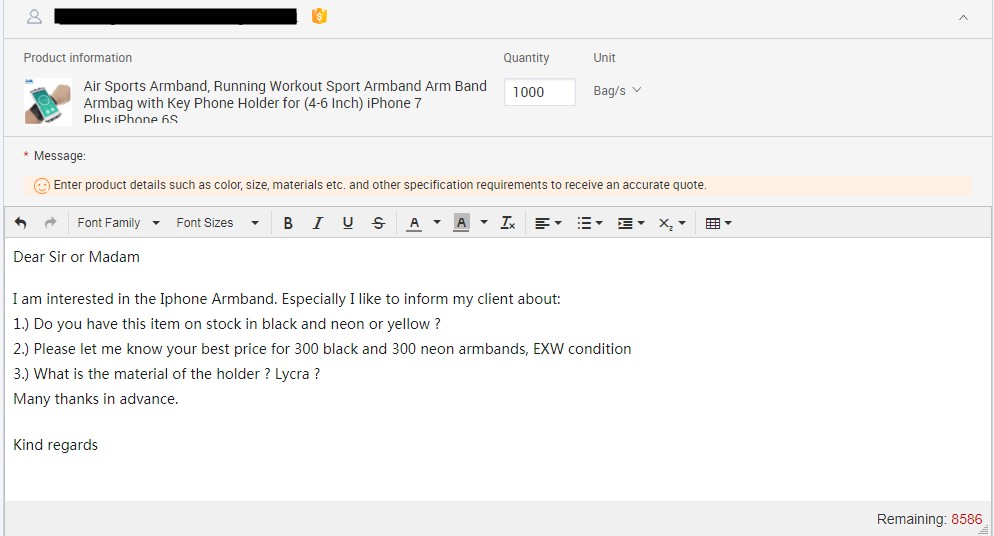}
	\caption{A typical product enquiry example on Alibaba.com} \label{jietu}
\end{figure}
Since the B2B procurement usually involves a large amount of money, the business will be a long process beginning with a product enquiry. Generally speaking, when customers are interested in some product, they will start a communication cycle with a seller by sending a product enquiry to the seller. In the product enquiry, customers will specify their requirements and ask questions about the product. Their requirements and questions usually refer to the most important attributes of the product. Fig.~\ref{jietu} shows an enquery example. Alibaba.com has accumulated tens of millions of product enquires, and we would like to leverage these information, in combination of the product knowledge graph we have, to figure out the most important attributes for each category of products.  

In our application scenario, the product knowledge graph is the existing knowledge graph and the enquiry data is the external textual data source. From now on, we will use our application scenario to explain the details of our proposed algorithm. 

We propose an unsupervised learning framework for extracting important product attributes from product enquiries. By calculating the semantic similarity between each enquiry sentence and each attribute of the product to which the enquiry corresponds to, we identify the product attributes that the customer cares about most. 

The attributes described in the enquiry may contain attribute names or attribute values or other expressions, for example, either the word ``color'' or a color instance word ``purple'' is mentioned. Therefore, when calculating the semantic similarity between enquiry sentences and product attributes, we need both attribute names and attribute values.  The same as any other knowledge graph, the product attributes in our knowledge graph we use contain noises and mistakes. We need to clean and normalize the attribute data before consuming it. We will introduce the detail of our data cleaning process in Section~\ref{subsec:datapreprocessing}.

\subsection{FastText Introduction}
FastText is a library created by the Facebook Research for efficient learning of word representations and sentence classification. Here, we just use the word representation functionality of it. 

FastText models morphology by considering subword units, and representing words by a sum of its character n-grams~\cite{bojanowski2016enriching}. In the original model the authors choose to use the binary logistic loss and the loss for a single instance is written as below:
\[
log\left(1+e^{-s\left(w_t,w_c\right)}\right)+\sum_{n\in \mathcal{N}_{t,c} }{log\left(1+e^{s\left(w_t,n\right)}\right)}
\]
By denoting the logistic loss function $\ell:x \rightarrow log\left(1+e^{-x}\right)$, the loss over a sentence is:
\[
\sum^{T}_{t=1}{\left[\sum_{c\in \mathcal{C}_t}{\ell\left(s\left(w_t,w_c\right)\right)}+\sum_{n\in \mathcal{N}_{t,c}}{\ell \left(-s\left(w_t,n\right)\right)}\right]}    
\]
The scoring function between a word $w$ and a context word $c$ is: 
\[
s\left(w,c\right) = \sum_{g\in \mathcal{G}_w}{\mathbf{z}_g^\top \mathbf{v}_c}
\]
In the above functions, $\mathcal{N}_{t,c}$ is a set of negative examples sampled from the vocabulary, $\mathcal{C}_t$ is the set of indices of words surrounding word $w_t$, $\mathcal{G}_w \subset \left\{1,\cdots ,G\right\}$ is the set of n-grams appearing in word $w$, $G$ is the size of the dictionary we have for n-grams, $\mathbf{z}_g$ is a vector representation to each n-gram $g$.

\subsubsection{Benefits over word2vec or glove}
Compared with word2vec or glove, FastText has following advantages:
\begin{itemize}
    \item It is able to cover rare words and out-of-vocabulary(OOV) words. Since the basic modeling units in FastText are ngrams, and both rare words and OOV ones can obtain efficient word representations from their composing ngrams. Word2vec and glove both fail to provide accurate vector representations for these words. In our application, the training data is written by end customers, and there are many misspellings which easily become OOV words. 
    \item Character n-grams embeddings tend to perform superior to word2vec and glove on smaller datasets.
    \item FastText is more efficient and its training is relatively fast.
\end{itemize}

\subsection{Matching}
 In this section, how to compute the matching between an enquiry sentence and a product attribute is explained in detail. Our explanation here is for a certain product category, and other categories are the same.
\begin{figure}
    \includegraphics[width=\textwidth]{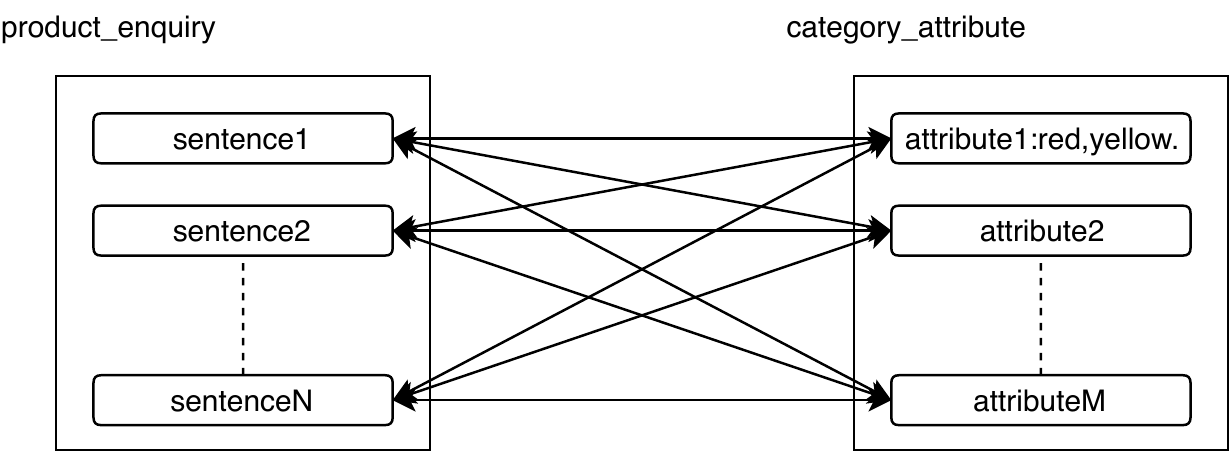}
    \caption{Each sentence obtained from the enquiry is scored against possible attributes under that category.} \label{xunpan}
\end{figure}

As you can see in Fig.~\ref{xunpan}, each sentence is compared with each attribute of a product category that the product belongs to. We now get a score between a sentence $s$ and an attribute $att$,
\[
vector\left(s\right) =\frac{\sum_{word \in s}v_{word}}{|s|}
\]
\[
vector\left(att\right) = \frac{\sum_{value\in V_{att}}{v_{value}}}{|V_{att}|}
\]
\[
score\left(s,att\right) = \cos \left(vector\left(s\right),vector\left(att\right)\right)
\]
where $V_{att}$ is all the possible values for this $att$, $v_{c}$ is the word vector for $c$.
According to this formula, we can get top two attributes whose scores are above the threshold $k$ for each sentence. We choose two attributes instead of one because there may be more than one attribute for each sentence. In addition, some sentences are greetings or self-introduction and do not contain the attribute information of the product, so we require that the score to be higher than a certain threshold.

\section{Experimentation}\label{sec:experiment}

\subsection{Data introduction}
For our knowledge graph data, entity(product) attributes can be roughly divided into clusters of transaction order specific ones and product specific ones, in this paper, we choose the product specific ones for further study. We also need to point out that we only focus on the recommended communication language on the Alibaba.com platform, which is \textbf{English}. 

To construct the evaluation dataset, top 14 categories are first chosen based on their business promotion features, and 3 millions typical products under each category were then chosen to form the attribute candidates. 
After preprocessing and basic filtering, top product specific attributes from the 14 different categories are chosen to be manually labeled by our annotators.
 
For each category, annotators each are asked to choose at most 10 important attributes from buyers perspective. After all annotators complete their annotations, attributes are then sorted according to the summed votes. 
In the end, 111 important attributes from the 14 categories are kept for final evaluation. 

Outside of the evaluation explained in this paper, we actually have performed the matching on more than 4,000 catetories covering more than 100 million products and more than 20 million enquires. Due to limited annotation resources, we can only sample a small numbered categories(14 here) to evaluate the proposed algorithm here. 
\subsection{Data preprocessing}\label{subsec:datapreprocessing}
The product enquiries and attributes data preprocessing is shown in Algorithm 1.
\floatname{algorithm}{Algorithm}
\renewcommand{\algorithmicrequire}{\textbf{Input:}}
\renewcommand{\algorithmicensure}{\textbf{Output:}}
\begin{algorithm}
	\caption{Data Preprocess Algorithm}
	\begin{algorithmic}[1]
		\REQUIRE $Product Enquiry$
		\ENSURE $Valid Sentences$: $VS$
		\STATE $VS \leftarrow \emptyset $
		\STATE $Remove Html Tag$: $Enquiry_{pt} \leftarrow Enquiry_{html}$
		\STATE Invalid $Enquiry_{pt}$ filter
		\STATE $S \leftarrow$  Split $Enquiry_{pt}$ to sentences
		\FOR{sentence $s$ in $S$}
		\STATE $Normalize(s)$
		\STATE $VS \leftarrow VS \cup s$
		\ENDFOR
		\STATE 
		\STATE \textbf{return $VS$}

	\end{algorithmic}  	
\end{algorithm}

Firstly, for every product enquiry, we convert the original html textual data into the plain text. Secondly we filter out the useless enquires, such as non-English enquires and spams. The regular expressions  and spam detection are used to detect non-English enquiries and spams respectively. Thirdly we get sentence list $S$ with spliting every enquiry  into sentences as described in section 2.2. Then for every sentence $s$ in $S$,  we need to do extra three processes: a)\emph{Spelling Correction}. b)\emph{Regular Measures and Numbers}. c)\emph{Stop Words Dropping}.
\begin{itemize}
	\item \emph{Spelling Correction}. Since quite a lot of the product enquires and self-filled attributes were misspelled, we have replaced the exact words by fuzzyfied search using Levenshtein distance. The method uses fuzzyfied search, only if the exact match is not found. Some attributes are actually the same, such as "type" and "product type", we merge these same attributes by judging whether the attributes are contained. 
	\item \emph{Regular Measures and Numbers}. Attributes of number type have their values composed of numbers and units, such as $128\times 300\times 350 cm^3$, $15.3 kg$, $220 V$, $50^\circ C$, etc. We replace all numbers (in any notation, e.g., floating point, scientific, arithmetical expression, etc.) with a unique token ($\sharp number\sharp$). For the same reason, each unit of measure is replaced with a corresponding token, eg., $cm^2$ is replaced with centimeter area.
	\item \emph{Stop Words Dropping}. Stop words appear to be of little value in the proposed matching algorithm. By removing the stop words we can focus on the important words instead. In our business scenario, we built a stop words list for foreign trade e-commerce.  
\end{itemize}

Finally, we get the valid sentences $VS$.

\subsection{Proposed method vs previous methods}

The existing co-occurrence methods do not suit our application scenario at all, since exact string matching is too strong a requirement and initial trial has shown its incompetency.  In stead we implemented an improved version of their method based on TextRank as our baseline. In addition, we also tested multiple semantic matching algorithms for comparison with our chosen method.
\begin{itemize}
\item TextRank: TextRank is a graph-based ranking model for text processing.\cite{mihalcea2004textrank} It is an unsupervised algorithm for keyword extraction. Since product attributes are usually the keywords in enquiries, we can compare these keywords with the category attributes and find the most important attributes. This method consists of three steps. The first step is to merge all enquiries under one category as an article. The second step is to extract the top 50 keywords for each category. The third step is to find the most important attributes by comparing top keywords with category attributes.  

\item Word2vec\cite{mikolov2013distributed}: 
We use the word vector trained by \cite{mikolov2013distributed} as the distributed representation of words. Then we get the enquiry sentence representation and category attribute representation. Finally we collect the statistics about the matched attributes of each category, and select the most frequent attributes under the same category.

\item GloVe\cite{pennington2014glove}: GloVe is a global log-bilinear regression model for the unsupervised learning of word representations, which utilizes the ratios of word-word co-occurrence probabilities. We use the GloVe method to train the distributed representation of words. And attribute selection procedure is the same as word2vec.
\end{itemize}
Proposed method: the detail of our proposed algorithm has been carefully explained in Section~\ref{sec:ourmethod}. There are several thresholds we need to pick in the experimentation setup. Based on trial and error analysis, we choose 0.75 as the sentence and attribute similarity threshold, which balances the precision and recall relatively well. In our application, due to product enquiry length limitation, customers usually don't refer to more than five attributes in their initial approach to the seller, we choose to keep 5 most important attributes for each category.

Evaluation is conducted by comparing the output of the systems with the manual annotated answers, and we calculate the precision and recall rate.

\[
Precision = \frac{\sum_{attr \in M_{a}~ \&~ attr \in S_{a}}1}{\sum_{ attr \in S_{a}}1}
\]
\[
Recall =\frac{\sum_{attr \in M_{a}~ \& ~attr \in S_{a}}1}{\sum_{attr \in M_{a}~}1}   
\]
where $M_{a}$ is the manually labeled attributes , $S_{a}$ is the detected important attributes.

Table 1 depicts the algorithm performance of each category and the overall average metrics among all categories for our approach and other methods. It can be observed that our proposed method  achieves the best performance. The average F1-measure of our approach is 0.47, while the average F1-measure values of “GloVe”, “word2vect” and "TextRank" are 0.46, 0.42 and 0.20 respectively. 

\begin{table}
	\renewcommand\arraystretch{1.2}
	\centering  
	\begin{threeparttable}
		\caption{Proposed method vs other methods metrics: precision, recall and F1-score.}  
		\begin{tabular}{l||ccc||ccc||ccc||ccc}  
			\toprule  
			match& \multicolumn{3}{c}{textrank}&\multicolumn{3}{c}{word2vec}& \multicolumn{3}{c}{glove}&\multicolumn{3}{c}{our approach}\cr  
			\cmidrule(lr){2-13}
			&P&R&F1&P&R&F1&P&R&F1&P&R&F1\cr  
			\midrule  
			Phone Bags        &  0.40  &  0.25  &  0.31&  0.60  &  0.38  &  0.46&  0.60  &  0.38  &  0.46 &  0.60  &  0.38  &  0.46\cr
			\hline
			Toys              &  0.40  &  0.22  &  0.29&  1.00  &  0.56  &  0.71&  1.00  &  0.56  &  0.71 &  1.00  &  0.56  &  0.71\cr
			\hline
			Books             &  0.60  &  0.50  &  0.55&  0.80  &  0.67  &  0.73&  0.80  &  0.67  &  0.73 &  0.80  &  0.67  &  0.73\cr
			\hline
			Handbags          &  0.40  &  0.25  &  0.31&  0.60  &  0.38  &  0.46&  1.00  &  0.62  &  0.77 &  1.00  &  0.62  &  0.77\cr
			\hline
			Earphones         &  0.20  &  0.04  &  0.07&  0.80  &  0.17  &  0.28&  0.80  &  0.17  &  0.28 &  0.80  &  0.17  &  0.28\cr
			\hline
			Traffic Light     &  0.20  &  0.07  &  0.11&  0.50  &  0.07  &  0.12&  0.60  &  0.21  &  0.32 &  0.60  &  0.21  &  0.32\cr
			\hline
			Bottles           &  0.20  &  0.07  &  0.11&  0.40  &  0.14  &  0.21&  0.60  &  0.21  &  0.32 &  0.60  &  0.21  &  0.32\cr
			\hline
			Mobile Phones     &  0.40  &  0.11  &  0.17&  1.00  &  0.22  &  0.36&  0.80  &  0.22  &  0.35 &  0.80  &  0.22  &  0.35\cr
			\hline
			Prefab Houses     &  0.00  &  0.00  &  0.00&  0.40  &  0.18  &  0.25&  0.40  &  0.18  &  0.25 &  0.40  &  0.18  &  0.25\cr
			\hline
			Stamps            &  0.00  &  0.00  &  0.00&  1.00  &  0.50  &  0.67&  0.75  &  0.50  &  0.60 &  1.00  &  0.50  &  0.67 \cr
			\hline
			Motorcycles       &  0.00  &  0.00  &  0.00&  0.60  &  0.20  &  0.30&  0.60  &  0.20  &  0.30 &  0.80  &  0.27  &  0.40\cr
			\hline
			Other Motor       &  0.80  &  0.25  &  0.38&  0.40  &  0.12  &  0.19&  0.60  &  0.19  &  0.29 &  0.60  &  0.19  &  0.29\cr
			\hline
			Elbow             &  0.40  &  0.29  &  0.33&  0.75  &  0.43  &  0.55&  0.80  &  0.57  &  0.67 &  0.75  &  0.43  &  0.55\cr
			\hline
			Power Banks       &  0.20  &  0.07  &  0.11&  0.60  &  0.21  &  0.32&  0.40  &  0.14  &  0.21 &  0.60  &  0.21  &  0.32\cr
			\hline
			average           &  0.30  &  0.15  &  0.20&  0.68  &  0.30  &  0.42&  0.70  &\bf{0.34}  &  0.46 &\bf{0.74}  &\bf{0.34}  &\bf{0.47}\cr
			\bottomrule
		\end{tabular} 
	\end{threeparttable}  
\end{table}

\subsection{Result Analysis}
In all our experiments, we find that FastText method outperforms other methods. By analyzing all results, we observe that semantic similarity based methods are more effective than the previous method which we implemented based on TextRank. This conclusion is understandable because lots of enquiries do not simply mention attribute words exactly, but some semantically related words are also used.

Evaluating FastText, GloVe and word2vec, we show that compared to other word representation learning algorithms, the FastText performs best. We sample and analyze the category attributes and find that many self-filled attributes contain misspellings. The FastText algorithm represents words by a sum of its character n-grams and it is much robust against problems like misspellings. In summary, FastText has greater advantages in dealing with natural language corpus usually with spelling mistakes. 

We also applied the detected attributes in the automatic enquiry generation task and we obtained significantly better generated enquiries compared to previous rigid templates. Due to space limitation, we skip the explanation and leave it for future publications.

\section{Conclusions and Future work }\label{sec:futurework}
In this paper, we proposed a new general method of identifying important attributes for entities from a knowledge graph. This is a relatively new task and our proposed method of using external textual data and performing semantic matching via word/sub-word embeddings obtained better result compared to other work of using naive string matching and counting.  In addition, we also successfully applied the detected important attributes in our real world application of smart composing. In summary, the method is extensible to any knowledge graph without attribute importance information and outperforms previous method. 

In future work, there are two major areas with potential of improving the detection accuracy. The first one is about sentence splitting. What we are trying to get is semantic cohesive unit, which can be used to match an attribute, and there might be more comprehensive method than the simple splitting by sentence ending punctuations. The second one is about improving the word embedding quality. We have implemented an in-house improved version of Fasttext, which is adapted to our data source. It is highly possible to use the improved word embedding on purpose of obtaining higher semantic matching precision.  As for the application, we will try to use more statistical models in the natural language generation part of the smart composing framework of consuming the detected important attributes. 



%
%
%
%
%
\bibliographystyle{splncs04}
\bibliography{mybibliography}

\begin{thebibliography}{10}
\providecommand{\url}[1]{\texttt{#1}}
\providecommand{\urlprefix}{URL }
\providecommand{\doi}[1]{https://doi.org/#1}

\bibitem{alientity}
Ali, E., Caputo, A., Lawless, S.: Entity attribute ranking using learning to
  rank

\bibitem{atzori2014ranking}
Atzori, M., Dessi, A.: Ranking dbpedia properties. In: WETICE Conference
  (WETICE), 2014 IEEE 23rd International. pp. 441--446. IEEE (2014)

\bibitem{bojanowski2016enriching}
Bojanowski, P., Grave, E., Joulin, A., Mikolov, T.: Enriching word vectors with
  subword information. arXiv preprint arXiv:1607.04606  (2016)

\bibitem{ding2005finding}
Ding, L., Pan, R., Finin, T., Joshi, A., Peng, Y., Kolari, P.: Finding and
  ranking knowledge on the semantic web. In: International Semantic Web
  Conference. pp. 156--170. Springer (2005)

\bibitem{ali2017attribute}
Esraa, A., Annalina, C., Séamus, L.: Attribute extraction and scoring: A
  probabilistic approach. In: KG4IR@SIGIR (CEUR Workshop Proceedings).
  vol.~1883, pp. 19--24. SIGIR (2017)

\bibitem{graves2008method}
Graves, A., Adali, S., Hendler, J.: A method to rank nodes in an rdf graph. In:
  Proceedings of the 2007 International Conference on Posters and
  Demonstrations-Volume 401. pp. 84--85. CEUR-WS. org (2008)

\bibitem{he2010xhrank}
He, X., Baker, M.: xhrank: Ranking entities on the semantic web. In: 9th
  international semantic web conference (ISWC2010) (2010)

\bibitem{hogan2006reconrank}
Hogan, A., Decker, S., Harth, A.: Reconrank: A scalable ranking method for
  semantic web data with context  (2006)

\bibitem{lee2013attribute}
Lee, T., Wang, Z., Wang, H., Hwang, S.w.: Attribute extraction and scoring: A
  probabilistic approach. In: Data Engineering (ICDE), 2013 IEEE 29th
  International Conference on. pp. 194--205. IEEE (2013)

\bibitem{lin15relationship}
Lin, Y., Liu, Z., Sun, M.: Modeling relation paths for representation learning
  of knowledge bases. CoRR  \textbf{abs/1506.00379} (2015),
  \url{http://arxiv.org/abs/1506.00379}

\bibitem{mihalcea2004textrank}
Mihalcea, R., Tarau, P.: Textrank: Bringing order into text. In: Proceedings of
  the 2004 conference on empirical methods in natural language processing
  (2004)

\bibitem{mikolov2013distributed}
Mikolov, T., Sutskever, I., Chen, K., Corrado, G.S., Dean, J.: Distributed
  representations of words and phrases and their compositionality. In: Advances
  in neural information processing systems. pp. 3111--3119 (2013)

\bibitem{mousavi2013ibminer}
Mousavi, H., Gao, S., Zaniolo, C.: Ibminer: A text mining tool for constructing
  and populating infobox databases and knowledge bases. Proceedings of the VLDB
  Endowment  \textbf{6}(12),  1330--1333 (2013)

\bibitem{pacsca2010role}
Pa{\c{s}}ca, M., Alfonseca, E., Robledo-Arnuncio, E., Martin-Brualla, R., Hall,
  K.: The role of query sessions in extracting instance attributes from web
  search queries. In: European Conference on Information Retrieval. pp. 62--74.
  Springer (2010)

\bibitem{pasca2007you}
Pasca, M., Van~Durme, B.: What you seek is what you get: Extraction of class
  attributes from query logs. In: IJCAI. vol.~7, pp. 2832--2837 (2007)

\bibitem{pacsca2007role}
Pa{\c{s}}ca, M., Van~Durme, B., Garera, N.: The role of documents vs. queries
  in extracting class attributes from text. In: Proceedings of the sixteenth
  ACM conference on Conference on information and knowledge management. pp.
  485--494. ACM (2007)

\bibitem{pennington2014glove}
Pennington, J., Socher, R., Manning, C.: Glove: Global vectors for word
  representation. In: Proceedings of the 2014 conference on empirical methods
  in natural language processing (EMNLP). pp. 1532--1543 (2014)

\bibitem{rai2012identifying}
Rai, R.: Identifying key product attributes and their importance levels from
  online customer reviews. In: ASME 2012 international design engineering
  technical conferences and computers and information in engineering
  conference. pp. 533--540. American Society of Mechanical Engineers (2012)

\bibitem{amit2012knowledgegraph}
Singhal, A.: Introducing the knowledge graph: things, not strings,
  \url{https://www.blog.google/products/search/introducing-knowledge-graph-things-not/}

\bibitem{vadrevu2016ranking}
Vadrevu, S., Tu, Y., Salvetti, F.: Ranking relevant attributes of entity in
  structured knowledge base (Jan~5 2016), uS Patent 9,229,988

\bibitem{2014wangembedding}
Wang, Z., Zhang, J., Feng, J., Chen, Z.: Knowledge graph embedding by
  translating on hyperplanes. In: Proceedings of the Twenty-Eighth AAAI
  Conference on Artificial Intelligence. pp. 1112--1119. AAAI'14, AAAI Press
  (2014), \url{http://dl.acm.org/citation.cfm?id=2893873.2894046}

\end{thebibliography}
\end{document}